\documentclass[journal]{IEEEtran}

\usepackage[pdftex]{graphicx}
\graphicspath{{../pdf/}{figures/}}
\DeclareGraphicsExtensions{.pdf,.jpeg,.png}

\usepackage{amssymb}
\usepackage{amsmath}

\usepackage{multirow}
\newcommand*{\kone}{ConvAsm1x1U}
\newcommand*{\kocl}{ConvOclDirectFwd1x1}
\newcommand*{\kbone}{ConvAsmBwdWrW1x1}
\newcommand*{\kbthree}{ConvAsmBwdWrW3x3}
\newcommand*{\kpp}{kernel parameter prediction}

\newcommand*{\sts}{Seq2Seq}
\newcommand*{\opcs}{output parameter constraint satisfaction}
\newcommand*{\oktp}{output kernel tuning parameters}

\begin{document}
    \title{Optimal Kernel Tuning Parameter Prediction using Deep Sequence Models}

\author{Khawir Mahmood, Jehandad Khan, Hammad Afzal
\thanks{K. Mahmood, H. Afzal are with the National University of Sciences and Technology, Islamabad, Pakistan.}
\thanks{J. Khan is with AMD Inc., Austin, Texas, USA}
}


\maketitle

\begin{abstract}
GPU kernels have come to the forefront of computing due to their utility in varied fields, from high-performance computing to machine learning. A typical GPU compute kernel is invoked millions, if not billions of times in a typical application, which makes their performance highly critical. Due to the unknown nature of the optimization surface, an exhaustive search is required to discover the global optimum, which is infeasible due to the possible exponential number of parameter combinations. In this work, we propose a methodology that uses deep sequence-to-sequence models to predict the optimal tuning parameters governing compute kernels. This work considers the prediction of kernel parameters as a sequence to the sequence translation problem, borrowing models from the Natural Language Processing (NLP) domain. Parameters describing the input, output and weight tensors are considered as the \emph{input language} to the model that emits the corresponding kernel parameters. In essence, the model \emph{translates} the problem parameter language to kernel parameter language. The core contributions of this work are: a) Proposing that a sequence to sequence model can accurately learn the performance dynamics of a GPU compute kernel b) A novel network architecture which predicts the kernel tuning parameters for GPU kernels, c) A constrained beam search which incorporates the physical limits of the GPU hardware as well as other expert knowledge reducing the search space. The proposed algorithm can achieve more than 90\% accuracy on various convolutional kernels in MIOpen, the AMD machine learning primitives library. As a result, the proposed technique can reduce the development time and compute resources required to tune unseen input configurations, resulting in shorter development cycles, reduced development costs, and better user experience.
\end{abstract}




\section{Introduction}
\label{sec:intro}

\IEEEPARstart {G}{eneral}-Purpose Computing on Graphics Processing Units (GPGPU) has enabled modern Machine Learning (ML) algorithms to perform at-par and, in some cases, superior to humans \cite{OptMemEffCNNGPU, MLQuantumChem, DeepMindAlphaGo}. 
ML libraries such as Tensorflow \cite{abadi2016tensorflow} and Pytorch \cite{paszke2017pytorch} convert the data flow graph into smaller operations, which are optimized to exploit the parallelism offered by a GPU. These operations take the form of compute-intensive routines called \emph{kernels}; performance optimized kernels require a deep understanding of the target hardware to extract the maximum possible performance and enable an algorithm to run at high efficiency. The design process creates a tight coupling between the kernel and the target hardware, making it challenging to innovate and adapt to changing requirements. 

The standard approach to enhance code porting and re-usability between different variants of the same hardware class is to parameterize the kernels and perform a parameter search on the target hardware to ascertain their optimum values to reach peak performance. In some cases, these kernel parameters are hand-tuned for a particular architecture by engineers with a deep understanding of the hardware architecture, resulting in a tedious procedure that is difficult to scale for production software.

For production software, search techniques discover {\oktp} wherein kernels with different parameter values are compiled and benchmarked to arrive at the optimal parameter values \cite{sim4:KernelTuner, sim5:CLTune}.
Various heuristics are employed to avoid performing an exhaustive search, but the cost in terms of computing resources and time remains significant. This approach makes it difficult to generalize to unseen input configurations, a critical feature for production software.

Predicting the optimum tuning parameters is a challenging task due to the non-trivial nature of operations involved (such as convolutions and batch normalization), hardware complexity, and code design. These factors manifest as a non-linear optimization surface that is discrete and inexpressible as an analytical model; furthermore, the combinatorial nature of the parameters makes it infeasible to traverse the search space exhaustively.
This paper proposes an ML-based approach using sequence models (Recurrent Neural Networks) for predicting these parameters. By incorporating a novel constraint satisfaction procedure, the proposed method ensures that the predicted parameters are always valid for the target architecture as well as improve performance by reducing the search space. Moreover, the Recurrent Neural Network (RNN) learns the underlying dynamics of the kernel and can generalize to unseen problem configurations.

The core contributions of this work are: a) Proposing that a sequence to sequence model can accurately learn the performance dynamics of a GPU compute kernel b) A novel network architecture which predicts the kernel tuning parameters for GPU kernels, c) A constrained beam search which incorporates the physical limits of the GPU hardware as well as other expert knowledge reducing the search space.

The paper's organization is as follows: Section \ref{sec:review} reviews existing kernel parameter tuning techniques, while Section \ref{sec:dev} details the mechanics of the proposed algorithm. Section \ref{sec:improv} further improves the {\sts} framework by incorporating a constrained beam search. Section \ref{sec:results} compares the accuracy of the proposed work against classical machine learning methods as well as contrasts the performance of the various proposed architectures, section \ref{sec:concl} concludes the paper.
\section{Prior Work}
\label{sec:review}

Prior art for kernel tuning consists of two main areas: model-based, and black-box approaches. Model-based approaches aim to develop an analytical model to inform the search process about the behavior of the target hardware. On the other hand, model-less methods tend to treat the problem as an optimization problem aimed to discover the optimum by using domain-specific variants of some known optimization algorithms. The following paragraphs highlight the salient works in these areas and contrast them with the proposed approach. While these approaches succeed to varying degrees in designing efficient heuristics to discover the optimum tuning parameters, the actual optimum may only be found by an exhaustive evaluation of the search space, as is typically the case with combinatorial optimization problems.

The lack of an underlying model in black-box optimization-based approaches enables these techniques on various hardware targets without modification. However, this also means that the algorithm needs to be run from the start for each program, implying results from one execution may not improve successive runs. On the other hand, model-based approaches suffer from the limitations of the models they rely on, which may be different for various architectures. Machine-Learning based approaches promise to strike a balance between the two by automating the model design process as well as the ability to generalize based on the supplied data. In the following paragraphs, we begin by reviewing prevalent work in model-free approaches to kernel tuning, followed by a few methods based on model-based techniques.

Papenhausen et al. \cite{sim1:CodingAnts} extend the ant colony optimization technique \cite{dorigo2006ant} to discover optimal tuning parameters for kernels generated as part of the PPCG compiler framework \cite{verdoolaege2013polyhedral}. By incorporating performance into the cost model and sharding the search space, the algorithm can efficiently search the parameter space. Workhoven et al. \cite{sim4:KernelTuner} present a framework that enables the tuning of various kernels with different optimization algorithms. Kernel tuner contributes implementations for various optimization algorithms such as Basin Hopping, Differential Evolution, and the Firefly algorithm \cite{yang2013firefly}. CLTune by Nugetern et al. \cite{sim5:CLTune} and AutoTuner by Bruel et al. \cite{sim8:AutoTuner} present frameworks that, while do not introduce new techniques, enable tuning for kernels using search strategies such as simulated annealing and particle swarm optimization.



In contrast to optimization-based approaches, model-based methods aim to develop analytical models to discover the ideal set of parameters for a given kernel and hardware combination. Analytical models have the advantage of being faster to evaluate (compared to searching a space); however, they may lead to some losses due to model approximations and mismatches. Lim et al. \cite{sim2:StatnPred} use static analysis as the driving force to collect information about a GPU program, augmenting it with dynamic data collected during program execution. An analytical model, after that, predicts the set of optimum parameters based on the collected data. \emph{RoofLine} models \cite{sim11:RoofLine} predict the performance limits of a kernel and identify whether a kernel is compute-bound or memory-bound. However, translating such information to concrete performance parameter recommendations might not always be possible. The model makes certain assumptions about program execution, which may or may not be accurate in all cases. For example, in specific compute-intensive kernels, it might be better to launch fewer wave-fronts (or warps) and have each thread perform an increased amount of work. Furthermore, each new device generation obsoletes older models, making them less ideal for production use.

Machine Learning (ML) methods discover the underlying model using execution data, which promises to strike a balance between searching and hand-tuned models. Bergstra et al. \cite{sim7:BoostedRegTree} employ a boosted regression tree to predict the execution time of a compute kernel; this allows a search algorithm to discard bad candidates and search in a smaller space quickly. Falch et al. \cite{sim10:PerfPort} adopt a similar approach; they use an Artificial Neural Network to directly prune the search space by randomly sampling the parameter space to train a neural network. SnowPack \cite{sim3:SnowPack} and StaTuner \cite{sim6:StaTuner} use support vector machines as the regression models to estimate the runtime of a kernel and avoid benchmarking kernels by launching them.

The approaches discussed above use the machine learning model as an estimator of the kernel runtime; in contrast, DeepTune \cite{sim15:DeepTune} adopts a method that is closest to work presented here. In DeepTune, a deep neural network learns features directly from the kernel code like Natural Language Processing (NLP) models, using them to predict the kernel performance parameters. The proposed work improves upon this work by removing the need to parse the source code and incorporate hardware constraints dictated by the hardware and the kernel design. Furthermore, the proposed algorithm can predict as many as eight performance parameters as compared to only one in DeepTune. 
\section{Problem Description}
\label{sec:prob}
Modern neural network models rely on various operations to form compute graphs which learn the target function, however, the most useful one, particularly for vision-based tasks is the convolution operation. The convolution operation accounts for a considerable chunk of the training time spent to learn its parameters. Therefore, improved performance parameters for convolutional layers would yield considerable dividends in terms of better performance. Due to their criticality in modern networks and tuning complexity, they are the ideal candidates for a performance parameter search study. 

\subsection{The convolution operation}
Convolution in machine learning refers to the traditional de-convolution operation in engineering, where the learning algorithm learns the weights of the kernel using the data it is presented \cite{lecun1995convolutional}. The important notion here is that learning happens over the weights of the convolutional kernel as opposed to traditional machine vision where the weights of the kernel are known \emph{apriori}. Therefore, the convolution operation takes two inputs; the image being convolved and the weights of the kernels and produces the convolved image as the output. Formally:

\begin{align}
    \label{eq:conv_basic}
    O & = I \circledast W
\end{align}

Where $O$ represents the output image and $I$ and $W$ represent the input image and the convolution weights, respectively. The convolution operation itself is represented as $\circledast$. Keep in mind, the notion of input and output changes in during back-propagation with $W$ and $I$ being outputs during the backward weights and backward data operations. However, both those can again be expressed as regular convolution and therefore are irrelevant to the present discussion. 

To improve computational efficiency and enhance throughput multiple input images are grouped together in \emph{batches} and convolved together. Therefore, $I$ and $W$ become four dimensional tensors instead of being two dimensional images with various channels (i.e., three dimensional tensors). Therefore, equation \ref{eq:conv_basic} may be written as

\begin{align}
    \label{eq:conv}
    O(n, k, h_o, w_o) & = I(n, c, h_i, w_i) \circledast W(k, c, y, x)
\end{align}

Where $O$, $I$ and $W$ are the output, input, and weight tensors respectively as before. The letters in parenthesis represent the dimensionality of each tensor with $n$ representing the batch-size, $h_o$ and $w_o$ represent the height and width of the output image, while $h_i$ and $w_i$ is the height and width of the input image. $k$ represents the number of channels in the output image, $c$ represents the number of channels in the input image, $y$ and $x$ is the height and width of the convolution filter, respectively.

The dimensions of a tensor collectively are known as the \emph{Tensor Descriptor} for a particular tensor. Therefore, the input tensor is described by the tensor descriptor containing $n,c, h_i, w_i$, collectively these parameters represent the feature space for our kernel performance parameter predictor. A computer kernel (whether on the GPU or a CPU) converts the input and weight hyper-volume to an output hyper-volume, with the amount of memory and number of computation required directly proportional to their collective hyper-volume.

A kernel developer employs various optimizations around memory access patterns, data layouts, memory hierarchy, access width etc. to compute the output volume as efficiently as possible. Collectively, these choices comprise the design space of the kernel author, with different points in this space resulting in different performance characteristics. The difficulty lies in choosing the ideal parameters for a given set of input tensor descriptors, since the relative volume of the input tensors deeply affects the choice of performance parameters to achieve optimum performance. The proposed algorithm predicts these algorithmic choices based on the various input tensor descriptors. The choice of convolution as the problem space is motivated by the complexity and importance of this operation in various machine learning scenarios.
\section{Method}
\label{sec:dev}

The proposed model is based on traditional language models that translate a sequence of token in one language to an equivalent sequence of words in another language. These models are called Sequence-to-Sequence models \cite{s2s:DBLP:journals/corr/SutskeverVL14}. One class of such models is called Encoder-Decoder models due to the division of the network architecture in two distinct networks connected by a hidden state \cite{MIT_DLbasics}. For the task at hand, such models perform remarkably better compared to traditional classification networks in the following ways: a) A single model can learn to predict all the tuning parameters of a kernel, thus learning the joint probability distribution over the target domain. b) Due to the recurrent nature of the networks, they are better able to learn and generalize the correlations between different parameters. However, there are certain disadvantages to this choice of architecture as well; for instance, due to the token-based nature of the input and the output, the network cannot learn to predict unseen ranges. 

The integral nature of the input, output, and weight tensor descriptors makes it tempting to use their numerical values to predict the integral tuning parameters. However, there are certain caveats associated with this thinking. If an estimator function learns to predict based on the integral value of the height of an image, the learned function would be a continuous one, making fractional image sizes valid -- which is not correct. Furthermore, the behavior of the kernel may change at specific values, such as when some hardware resource is exhausted, which again is difficult to learn in a continuous setting. Therefore, in practice, the authors observed that encoding the tensor descriptors as well as the predicted performance parameters as one-hot encoded tokens yielded better results, as evidenced by the results section. In contrast to typical language processing tasks that embed the input sequence of tokens in some space \cite{DBLP:journals/corr/MikolovSCCD13}, \cite{pennington2014glove}, this work performs an end-to-end conversion between input tokens representing characteristics of the convolution operation and tokens representing performance parameters. 
\subsection{The Basic Encoder-Decoder Model}
\label{dev:edm}
In their seminal work Sutskever, et al. \cite{s2s:DBLP:journals/corr/SutskeverVL14} used the RNN architecture to outperform state-of-the-art on sequence-to-sequence prediction tasks. The strength of the model lies in its ability to train a single end-to-end model on the source and target sequences and to handle variable-length input and output sequences. The encoders and decoders comprise of Long Short-Term Memory(LSTM) \cite{s2s:DBLP:journals/corr/SutskeverVL14} or Gated Recurrent Unit (GRU) \cite{DBLP:journals/corr/ChoMGBSB14} cells. 

\begin{figure}[ht]
	\centering
	\includegraphics[clip=true, width=0.46\textwidth]{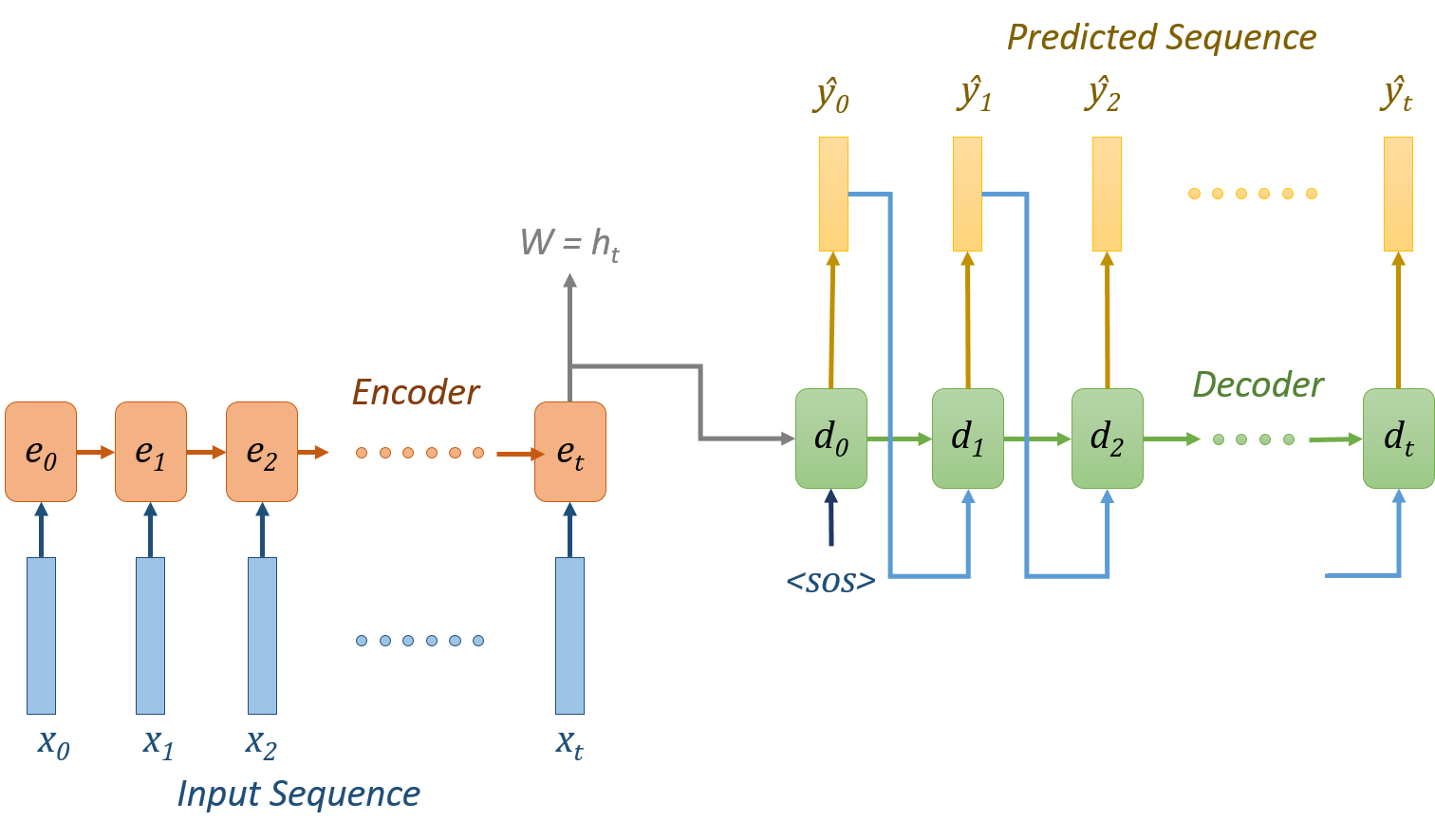}
	\caption{Encoder-Decoder Architecture}
	\label{fig:4-4_encoder-decoder}
\end{figure}

Figure~\ref{fig:4-4_encoder-decoder} shows the encoder-decoder model unrolled in the time-dimension; the encoder and decoder network itself consists of only one cell. For each sample, the input sequence $[x_{0},x_{1}, ... x_{t}]$ is fed to the encoder in consecutive time-steps $ 0 \rightarrow t$. The encoder updates its internal state $e$ at each time step, i.e., $ e_{0} \rightarrow e_{1} \rightarrow .... e_{t} $ using information from the input at that time step $i$ while preserving the information it has already learned from previous time-steps $0 \rightarrow i-1$. Thus, at final time-step \emph{t}, the whole input sequence is represented by the internal state $ e_{t}$ also known as thought vector $w$.

Based on $ e_{t}$, at each time step $t$, the decoder generates the output sequence $[y\hat{}_{0},y\hat{}_{1}, ... y\hat{}_{t}]$ as a probability distribution over the output parameter range. Note that, at each time-step after $t=0$, the predicted value from previous time step $y\hat{}_{t-1}$ is also used for making the prediction, i.e., updating the decoder's internal state $d_t$.

\subsection{Attention}
\label{dev:attn}
The Encoder-Decoder model relies on a single fixed-length \emph{thought vector} to represent the complete source sequence while discarding the useful information in the intermediate encoder states, resulting in inefficient handling of longer sequences. Augmenting the Encoder-Decoder networks with an \emph{Attention} mechanism \cite{bahdanau2014neural,DBLP:journals/corr/LuongPM15}  enables the decoder to focus on different sub-sequences during each step, overcoming the limitations discussed above.

\begin{figure}[ht]
	\centering
	\includegraphics[clip=true, width=0.46\textwidth]{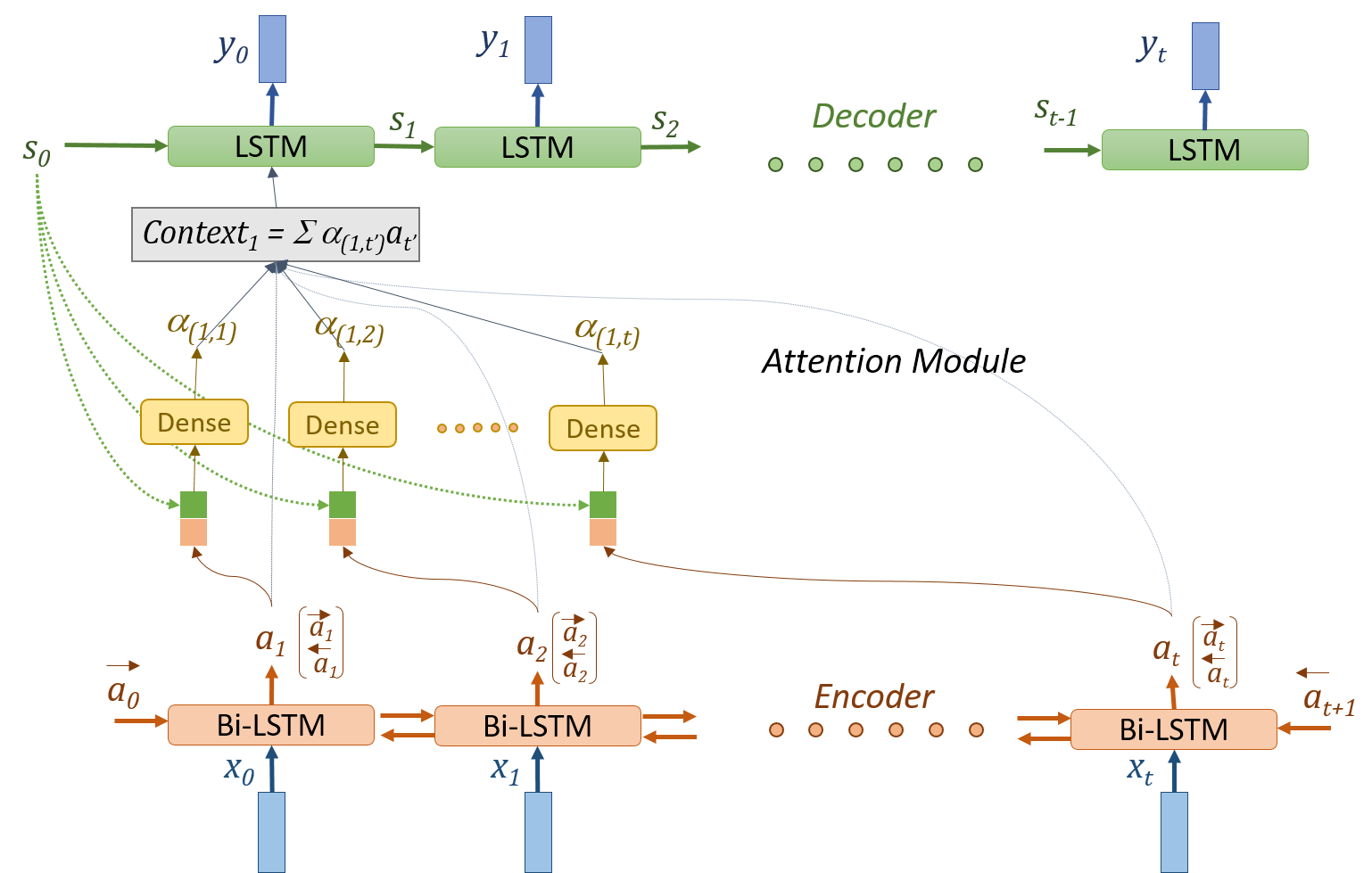}
	\caption{Proposed Attention Model}
	\label{fig:4-6_attention2}
\end{figure}

This work proposes a modified version of \emph{Bahdanau-attention} \cite{bahdanau2014neural} by not reusing intermediate outputs at each time-step, likely because of independent output tokens. The model shown in Figure~\ref{fig:4-6_attention2} consists of a bi-directional encoder LSTM called \emph{pre-attention bi-LSTM}, an \emph{attention mechanism}, and a decoder LSTM called \emph{post-attention LSTM}.

For each sample, the input sequence $[x_{0},x_{1}, ... x_{t}]$ is fed to the pre-attention bi-LSTM in consecutive time-steps $0 \rightarrow t$. For each input $x_i$, the pre-attention bi-LSTM outputs the forward $\overrightarrow{a_{i}}$ and backward $\overleftarrow{a_i}$ activations (internal states), concatenated together form $a_{i}$. 

\begin{equation}
a_{i} = \{\overrightarrow{a_{i}},\overleftarrow{a_i}\} 
\end{equation}

At each time-step $t$ the attention weights $\alpha_{(i,t')}$ over all input time-steps $t'$ are calculated using the previous post-attention LSTM internal state $s_{i-1}$ and concatenated activations $a_{i}$ of the pre-attention bi-LSTM. A Neural Network consisting of a single \emph{Dense} layer calculates the weights $d$ which are then passed through a \emph{Softmax} layer to get a probability distribution of attention weights $\alpha$.

\begin{equation}
d_{(i,t')} = \textup{Dense}(s_{i-1},a_{t'})
\alpha_{(i,t')}=\frac{\exp(d_{(i,t')})}{\sum_{k=1}^{t} \exp(d_{(i,t')})}
\end{equation}

The attention weights $\alpha_{(i,t')}$ are combined together to form the Context Vector $C_{i}$ which is calculated by taking a sum of the attention weights $\alpha_{(i,t')}$ weighted by the respective input activations $a_{t'}$.

\begin{equation}
C_{i} = \sum_{j=1}^{t}\alpha_{(i,t')}a_{t'}
\end{equation}

The Context Vector $C_{i}$ is used by the post-attention LSTM to sequentially predict $y_{i}$ as a probability distribution over the output parameter range.


\subsection{Adding Convolutions - Hybrid Model}
\label{dev:hybrid}
While the attention module, coupled with bi-directional LSTMs capture long term dependencies. However, Convolutional Neural Networks (CNN) are better adept at capturing spatial relationships in data and may be computed more efficiently than RNNs \cite{DBLP:journals/corr/KalchbrennerB13,NIPS2015_5782}. Popular due to their success in image-based classification tasks, convolutional layers have been used successfully in many other applications such as image captioning \cite{You2016CVPR}, change detection \cite{luo2017revisit}, and video activity recognition \cite{ZWen2017InfoProp,DBLP:journals/corr/abs-1808-03867,Gehring:2017:CSS:3305381.3305510}.

\begin{figure}[ht]
	\centering
	\includegraphics[clip=true, width=0.46\textwidth]{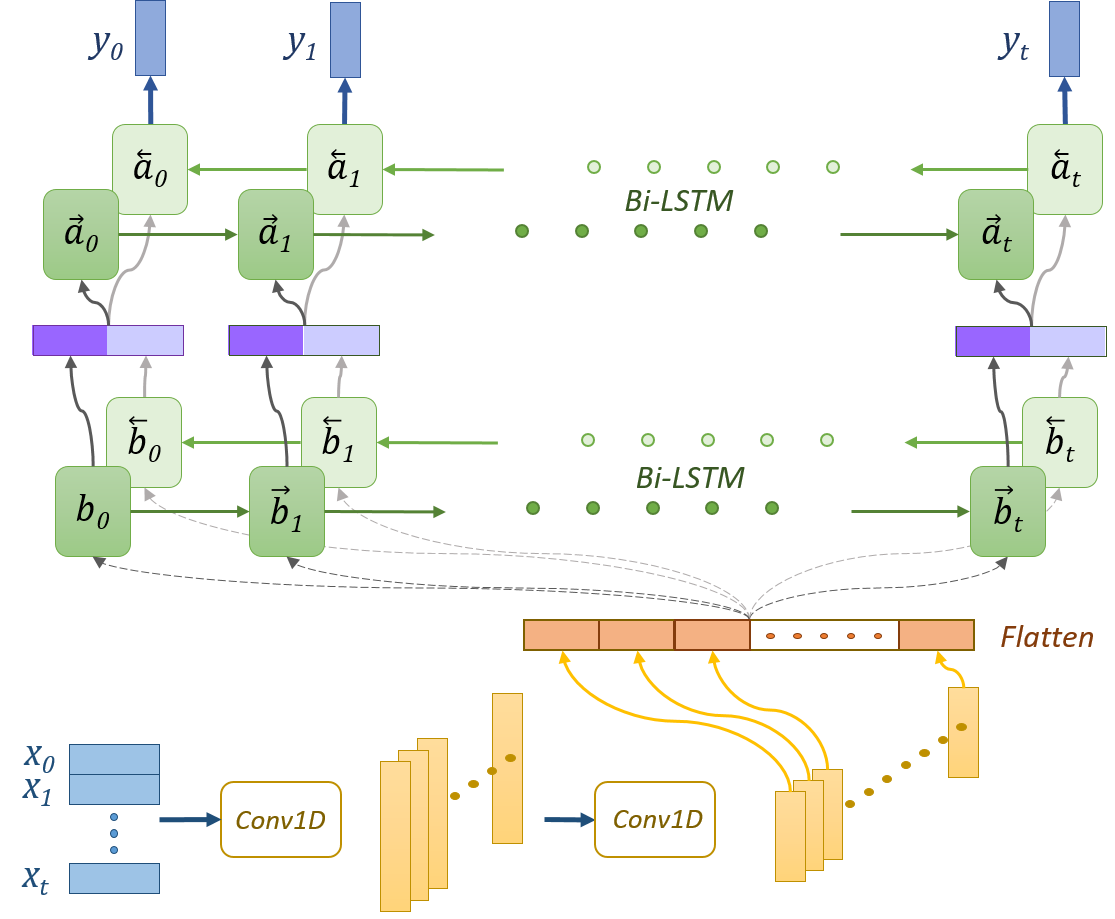}
	\caption{Conv1D encoder and Bi-LSTM decoder architecture}
	\label{fig:4-8_hybrid}
\end{figure}

In this spirit, the authors implemented a hybrid model shown in Figure~\ref{fig:4-8_hybrid} that uses convolutional (Conv) layers as the encoder followed by bi-LSTMs for the decoder.

The complete input sequence passes through multiple \emph{1-dimensional Convolutional layers} ($Conv1D$).  Each layer consists of various filters $f$, of size $k$ that perform convolutions with stride $s$. On an input sequence of length $t$, a $Conv1D$ forms $f$ representations of the data, each of size $o = \frac{t-k}{s}+1 $. Traditionally each convolutional layer is followed by a pooling layer. However, in the case of {\kpp}, adding pooling layers reduces model performance, likely due to the reduced size of the feature space, in contrast to image processing, where images have a large number of pixels. The output from the last $Conv1D$ layer is flattened to form a 1-dimensional vector $d= f \times o$, used to predict the complete output sequence. The decoder consists of multiple Bi-LSTM layers that sequentially predict the output sequence based on the flattened input from the Conv Encoder. The Bi-LSTM layer helps map interdependencies in the output sequence. Finally, a Dense layer is used with Softmax activation to create the output sequence likelihood with the highest value corresponding to the predicted output.

\begin{figure}[ht]
	\centering
	\includegraphics[clip=true, width=0.46\textwidth]{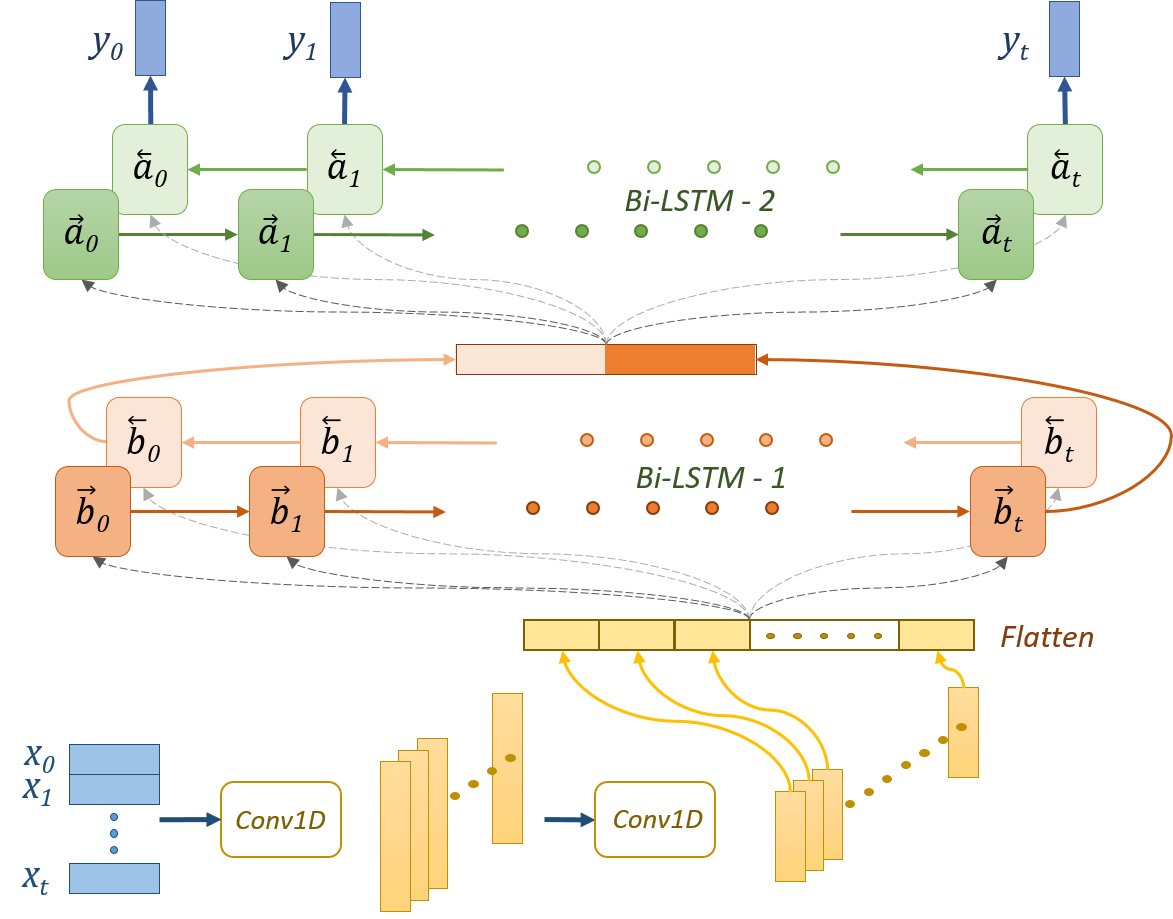}
	\caption{Conv1D encoder and layered bi-LSTM decoder architecture}
	\label{fig:4-9_hybrid-layered}
\end{figure}

Figure~\ref{fig:4-9_hybrid-layered} shows a variation of the Hybrid Model, where only the last states (forward $\overrightarrow{b_{t}}$ and reverse $\overleftarrow{b_{0}}$) from bi-LSTM-1 are passed on to bi-LSTM-2 removing false temporal dependencies formed amongst predicted output sequences when using multiple bi-LSTM layers.

\section{Constrained Beam Search}
\label{sec:improv}
\begin{figure}[ht]
	\centering
	\includegraphics[clip=true, width=0.46\textwidth]{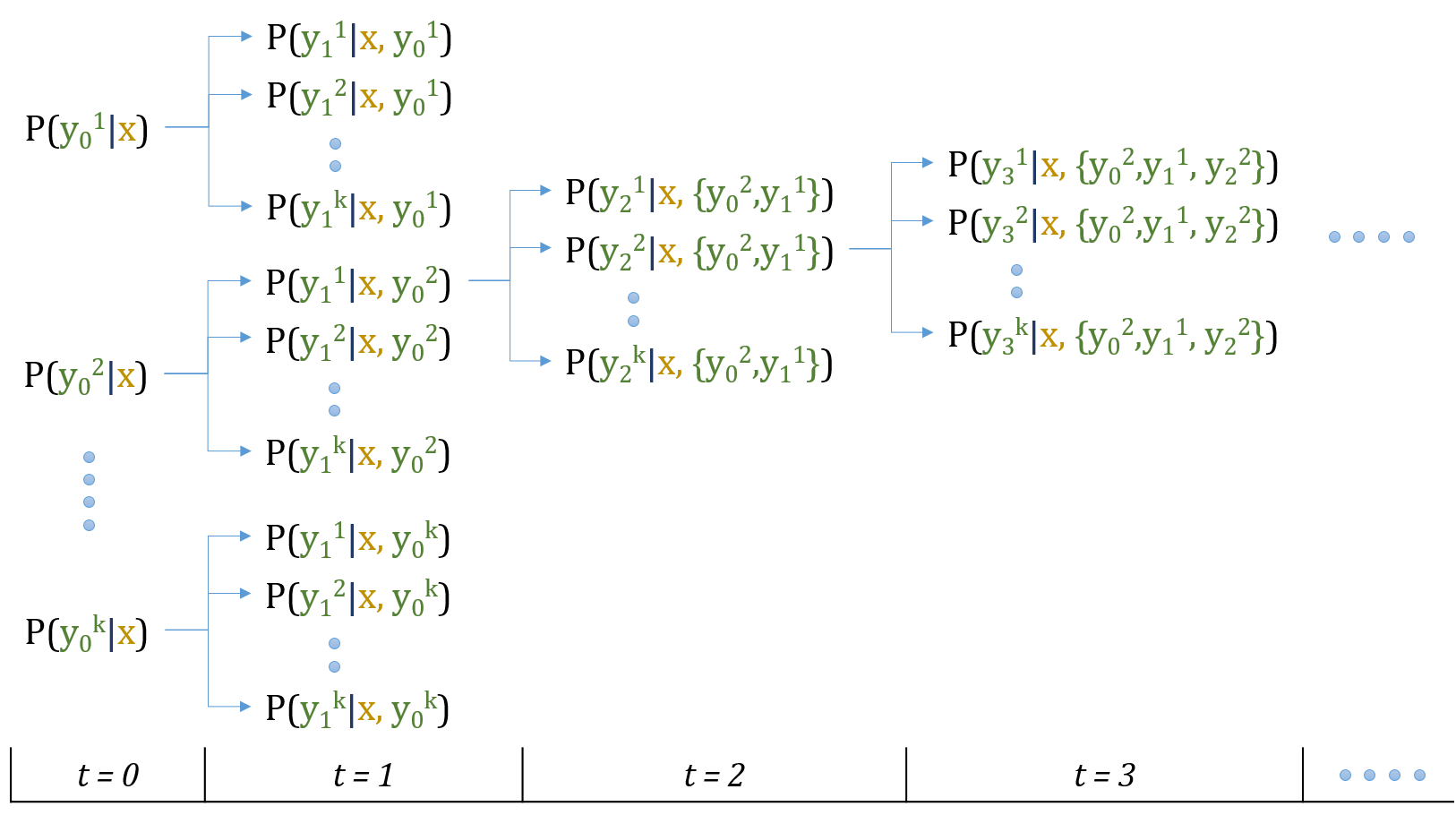}
	\caption{Beam Search For selecting $k$ values with highest probabilities at each step}
	\label{fig:5-1_beam-search}
\end{figure}

The decoder of an RNN computes the token probability at the time $i$ conditioned on the hidden vector state (post attention) as well as all output tokens until time $t-1$, implying that an erroneous prediction at any time may result in an incorrect prediction. This problem may be mitigated using a technique known as \emph{Beam Search} and is commonly used to improve prediction performance in case of sequence to sequence models \cite{DBLP:journals/corr/FreitagA17, DBLP:journals/corr/WisemanR16}. Instead of only propagating the word with the highest probability at each time step, the method propagates $k$ subsequences with the highest likelihood at each time step as shown in Figure \ref{fig:5-1_beam-search}. The parameter $k$ is known as \emph{Beam Width}. Incorporating beam search in the models presented above resulted in considerable gains in prediction accuracy, as discussed in detail below.

\begin{figure}[ht]
	\centering
	\includegraphics[clip=true, width=0.46\textwidth]{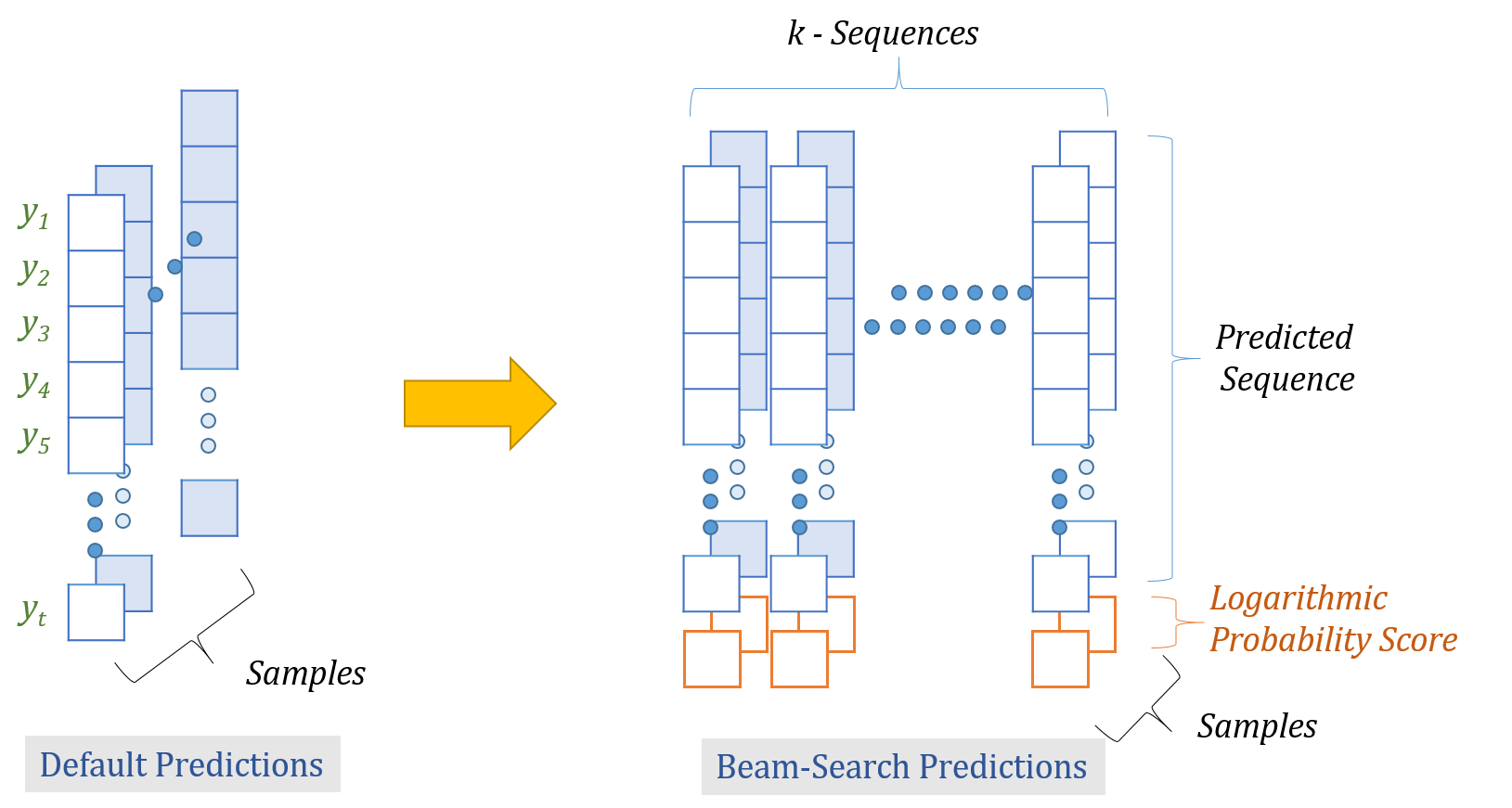}
	\caption{Change in shape of output kernel parameter sequence after applying Beam Search}
	\label{fig:5-2_beam-search-op}
\end{figure}

Figure \ref{fig:5-2_beam-search-op} depicts how beam search fits in the proposed architecture, where $y_{0 \rightarrowtail t}$ denotes time-steps or output parameters, and $y^{0 \rightarrowtail k}$ denotes the top $k$ choices. Thus, at each time-step $i$, $k$ best choices are kept based on the complete input sequence $x$ and the values leading up to the current time-step $y_{0 \rightarrowtail t-1}$.

\subsection{Applying Output Parameter Constraint Satisfaction}
\begin{figure}[ht]
	\centering
	\includegraphics[clip=true, width=0.47\textwidth]{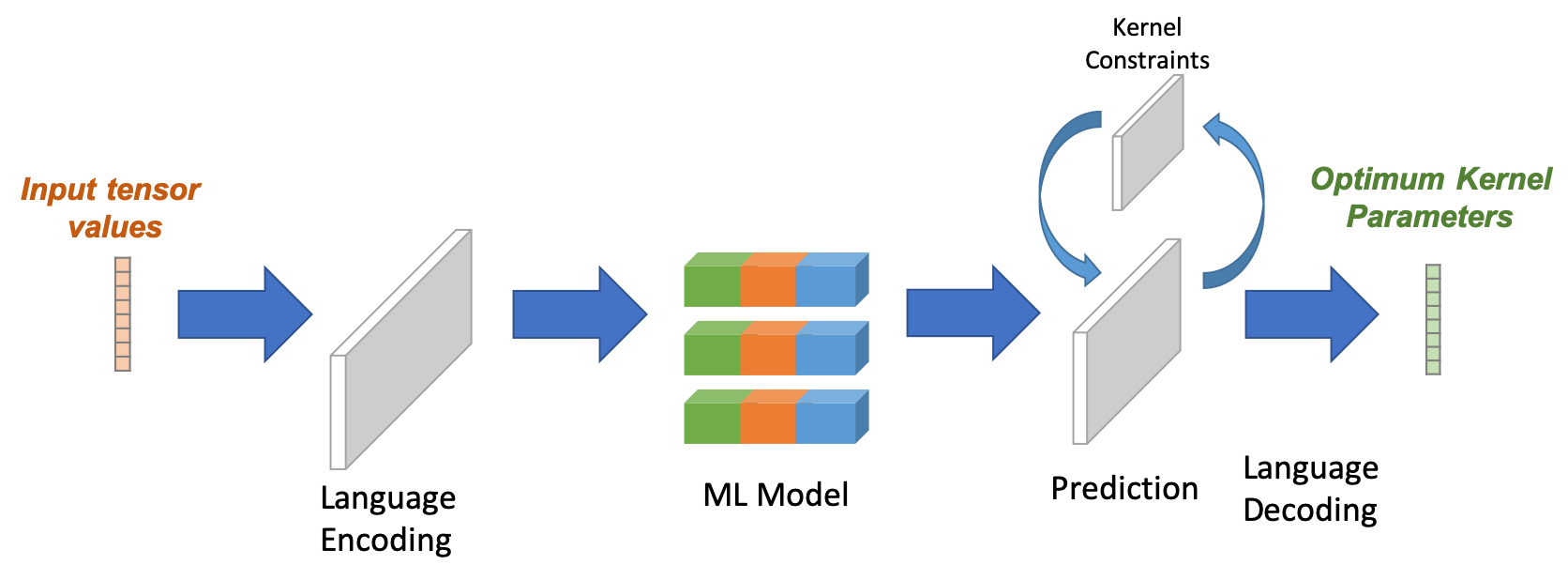}
	\caption{The overall flow of Improved Hybrid \emph{\sts} Framework showing language encoding, prediction, beam-search, constraint-satisfaction, and language decoding pipeline}
	\label{fig:5-3_constraints}
\end{figure}

The performance parameters for a compute kernel translate to the number of used hardware resources, such as the number of registers, amount of shared memory, types of instructions, etc. These resources have physical limits which, if violated, may either render a kernel invalid or result in drastic performance cliffs. This domain knowledge is invaluable in terms of reducing the search space for the predicted optimal tuning parameter. While the notion of constraining the output sequence in language translation models is not useful, for the case at hand presents a unique opportunity to improve prediction performance.  
These constraints are in the form of Boolean predicate functions that take the tensor descriptor and the predicted sub-sequence as inputs and results in the feasibility of the predicted parameter combination. Discarding the candidates that violate the output parameter constraints, considerably improves accuracy. Therefore, at each time-step the $k$ candidate sub-sequences are sure to be valid for the kernel in question, Figure~\ref{fig:5-3_constraints} depicts this process in more detail.

\section{Results and Analysis}
\label{sec:results}
Multiple kernels from MIOpen \cite{miopen} were selected as candidates, to ascertain the efficacy of the proposed neural architecture. MIOpen is AMD's open-source library for high-performance machine learning primitives supporting different operations such as convolution, batchnorm, and RNNs. MIOpen ships with optimal performance parameters for these kernels for various convolution configurations, these input parameters were processed according to the techniques discussed earlier. Table \ref{tab:3-3_outputs} outlines the selected kernels as well as the number and range of their performance parameters; the total number of possible parameters is vast, making it a formidable task to predict these parameters accurately.

\begin{table}[ht]
	\caption{Kernel Output Parameters and Ranges}
	\label{tab:3-3_outputs}
	\centering
	\begin{tabular}{l|l c} 
		\textbf{Kernel} & \textbf{Parameters} & \textbf{Range/Values} \\ \hline
		\multirow{8}{*}{\textbf{\kone}} & \textbf{read\_size} & 1-4 \\
		& \textbf{k\_mult} & 1,4,8,16,32 \\
		& \textbf{chunks\_per\_wave} & 1-16 \\
		& \textbf{chunk\_size} & 1,2,4,8,16,32,64 \\
		& \textbf{n\_mult} & 1-8 \\
		& \textbf{c\_mult} & 1,2,4,8,16,32 \\
		& \textbf{waves\_c\_in\_group} & 1-8 \\
		& \textbf{waves\_k\_in\_group} & 1,2,4,8 \\ \hline
		\multirow{9}{*}{\textbf{\kocl}} & \textbf{grp\_tile1} & $2^{0-4}$ \\
		& \textbf{grp\_tile0} & $2^{0-8}$ \\
		& \textbf{in\_tile1} & $2^{0-5}$ \\
		& \textbf{in\_tile\_0} & $2^{0-5}$ \\
		& \textbf{out\_pix\_tile1} & 0,1 \\
		& \textbf{out\_pix\_tile0} & 0,1,2,4 \\
		& \textbf{n\_out\_pix\_tiles} & $2^{0-6}$ \\
		& \textbf{n\_in\_data\_tiles} & $2^{0-11}$ \\
		& \textbf{n\_stacks} & 0,1 \\ \hline
		\multirow{10}{*}{\textbf{\kbone}} & \textbf{read\_size} & 1-4 \\
		& \textbf{c\_per\_gpr} & 1,2,4,8,16 \\
		& \textbf{c\_mult} & 1,2,4,8,16 \\
		& \textbf{k\_per\_gpr} & 1,2,4,8,16 \\
		& \textbf{k\_mult} & 1,2,4,8,16 \\
		& \textbf{n\_per\_gpr} & 1,2,4,8,16 \\
		& \textbf{n\_part\_cnt} & 1-8 \\
		& \textbf{chunk\_size} & 1,2,4,8,16 \\
		& \textbf{short\_store} & 0,1 \\
		& \textbf{data\_prefetch} & 0-4 \\ \hline
		\multirow{6}{*}{\textbf{\kbthree}} & \textbf{limit\_wave\_cnt} &  0-9\\
		& \textbf{reverse\_inout} & 0,1 \\
		& \textbf{chunk\_size} & 8,16\\
		& \textbf{k\_per\_wave} & 1,2,4,8\\
		& \textbf{pipe\_lines\_depth} & 1-16\\
		& \textbf{n\_per\_group} & 1-8\\ \hline
	\end{tabular}
\end{table}
%

To better assess the performance of various techniques, it is useful to define the following metrics:

The \emph{mean of the prediction accuracy} of each output parameter is defined as

\begin{equation*}
Average Accuracy = {\frac{1}{n}}\sum _{i=1}^{n}(\frac{C_{i}}{S_{Test}} * 100)
\end{equation*}

Where $n$ is the total number of output parameters for the kernel, $C_{i}$ is the number of correct predictions of parameter $i$ and $S_{Test}$ is the total number of test samples. The percentage of \emph{perfect predictions} in the test sample, where the complete predicted output parameter set matches the optimum parameter set for the test sample is given by:-

\begin{equation*}
Perfect Prediction = \frac{count(p_{0 \rightarrow t} = a_{0 \rightarrow t})}{S_{Test}}\times 100
\end{equation*}

Where $p_{0 \rightarrow t}$ is the predicted output parameter/sequence set, $a_{0 \rightarrow t}$ is the actual output parameter/sequence set, and $S_{Test}$ is the total number of test samples. This metric quantifies a joint prediction of the parameter sequence. Therefore, it is a more useful metric for evaluating model performance since the key objective is to predict the complete set of optimum kernel tuning parameters.

Classical ML techniques such as regression trees, clustering, and multi-layer perceptrons set a baseline for prediction performance. This exercise provided insight into the data and assurance that no trivially discoverable patterns were present. Results from this exercise are presented in Table \ref{tab:3-10_g-summary}.

\begin{table*}
	\caption{Average Accuracy {\kpp} benchmarks using Classic ML techniques}
	\label{tab:3-10_g-summary}
	\centering
	\begin{tabular}{|lcc|ccccc|c|c|}
		\hline
		\textbf{Kernel} & \textbf{Precision} & \textbf{Samples} & \textbf{D.Tree} & \textbf{GNB} & \textbf{k-NN} & \textbf{SVC} & \textbf{RFC} & \textbf{Average} & \textbf{Best} \\ \hline
		\multirow{2}{*}{\textbf{\kone}} & \textbf{Full} & \textbf{7,038} & 61.52 & 48.45 & 52.18 & 52.16 & {\textbf{64.80}} & 54.60 & 64.80 \\
		& \textbf{Half} & \textbf{3,291} & 78.25 & 58.15 & 62.53 & 61.14 & {\textbf{80.34}} & 66.43 & 80.34 \\ \hline
		\multirow{2}{*}{\textbf{\kocl}} & \textbf{Full} & \textbf{21,124} & 77.58 & 68.41 & 68.77 & 69.32 & {\textbf{79.02}} & 71.92 & 79.02 \\
		& \textbf{Half} & \textbf{3,349} & 81.49 & 72.87 & 76.58 & 76.35 & {\textbf{82.19}} & 77.06 & 82.19 \\ \hline
		\multirow{2}{*}{\textbf{\kbone}} & \textbf{Full} & \textbf{5,100} & 71.00 & 66.00 & 70.71 & 71.43 & {\textbf{74.96}} & 70.02 & 74.96 \\
		& \textbf{Half} & \textbf{1,606} & 73.04 & 61.37 & 66.46 & 66.09 & {\textbf{75.34}} & 67.28 & 75.34 \\ \hline
		\multirow{2}{*}{\textbf{\kbthree}} & \textbf{Full} & \textbf{698} & 60.63 & 47.10 & 52.66 & 55.07 & {\textbf{65.70}} & 54.71 & 65.70 \\
		& \textbf{Half} & \textbf{447} & 67.04 & 51.11 & 60.37 & 68.15 & {\textbf{72.22}} & 61.67 & 72.22\\ \hline
	\end{tabular}
\end{table*}

Table \ref{tab:4-2_results} details the performance metrics of the models described in Section \ref{sec:dev} for different kernels and precisions. Sequential models consistently perform better than the classic ML techniques as measured by the Average Accuracy metric. The hybrid layered decoder model described in Section \ref{sec:dev} outperforms all other variants in both average and perfect predictions.

\begin{table*}[ht]
	\caption{Average Accuracy and Perfect Prediction Results of Seq2Seq Model Variations}
	\label{tab:4-2_results}
	\centering
	\begin{tabular}{|lc|cc|cc|cc|cc|cc|}
		\hline
		\multirow{2}{*}{\textbf{Kernel}} & \multirow{2}{*}{\textbf{Precision}} & \multicolumn{2}{c|}{\textbf{Enc-Dec}} & \multicolumn{2}{c|}{\textbf{Attn}} & \multicolumn{2}{c|}{\textbf{Attn-2}} & \multicolumn{2}{c|}{\textbf{Hybrid}} & \multicolumn{2}{c|}{\textbf{Hybrid-2}} \\
		&  & \textbf{Avg} & \textbf{Pft} & \textbf{Avg} & \textbf{Pft} & \textbf{Avg} & \textbf{Pft} & \textbf{Avg} & \textbf{Pft} & \textbf{Avg} & \textbf{Pft} \\ \hline
		\multirow{2}{*}{\textbf{\kone}} & \textbf{Full} & 55.27 & 4.26 & 55.47 & 2.13 & 56.71 & 1.85 & {\textbf{63.25}} & {\textbf{6.96}} & 62.59 & 6.53 \\
		& \textbf{Half} & 79.85 & 31.21 & 80.57 & 27.27 & 80.87 & 30.00 & 82.16 & 31.21 & {\textbf{82.23}} & {\textbf{31.52}} \\ \hline
		\multirow{2}{*}{\textbf{\kocl}} & \textbf{Full} & 82.84 & 15.76 & 82.33 & 15.05 & 82.78 & 16.75 & 82.63 & 15.52 & {\textbf{83.64}} & {\textbf{18.93}} \\
		& \textbf{Half} & 87.06 & 30.15 & 86.07 & 25.37 & 85.57 & 26.27 & {\textbf{87.10}} & 29.25 & 86.73 & {\textbf{31.34}} \\ \hline
		\multirow{2}{*}{\textbf{\kbone}} & \textbf{Full} & 75.39 & 8.24 & 74.00 & 4.31 & 73.63 & 4.51 & {\textbf{75.96}} & {\textbf{10.39}} & 75.78 & 10.00 \\
		& \textbf{Half} & 77.08 & 12.53 & 75.52 & 11.18 & 74.72 & 12.42 & 76.58 & 8.07 & {\textbf{77.76}} & {\textbf{13.66}} \\ \hline
		\multirow{2}{*}{\textbf{\kbthree}} & \textbf{Full} & 73.43 & 11.59 & 65.46 & 7.25 & 72.46 & 11.94 & 75.24 & 12.86 & {\textbf{75.48}} & {\textbf{15.43}} \\
		& \textbf{Half} & 76.67 & 17.78 & 77.41 & 13.33 & 75.93 & 11.11 & 77.78 & 17.78 & {\textbf{79.26}} & {\textbf{22.22}} \\ \hline
		\multicolumn{12}{c}{\footnotesize Avg: Average Accuracy, Pft: Prefect Prediction, Enc-Dec: Encoder Decoder, Attn: Attention} \\
		\multicolumn{12}{c}{\footnotesize Attn-2: Modified Attention, Hybrid-2: Hybrid Layered Decoder Model}
	\end{tabular}
\end{table*}

Augmenting the Hybrid {\sts} Layered Decoder Model with a constrained beam search ensures that a valid set of output parameter sequences is produced and improves results prediction metrics, as shown in Table \ref{tab:5-1_results}. Although it increases the number of outputs from a single output kernel parameter set to $k$ sets (where $k$ is the beam width), compared to empirical optimization techniques, even keeping $k=100$ has practical significance due to an increase in Perfect Prediction metric.

\begin{table*}[ht]
    \caption{Results of Hybrid {\sts} Framework with {\opcs} fused Beam Search}
    \label{tab:5-1_results}
    \centering
    \begin{tabular}{|lc|c|cccc|cccc|}
    	\hline
    	\multirow{2}{*}{\textbf{Kernel}} & \multirow{2}{*}{\textbf{Precision}} & \multirow{2}{*}{\textbf{Metric}} & \multicolumn{4}{c|}{\textbf{Hybrid-2 with Beam Search}} & \multicolumn{4}{c|}{\textbf{Hybrid-2 with OPCS-BS}} \\
    	&  &  & \textbf{k=10} & \textbf{k=30} & \textbf{k=50} & \textbf{k=100} & \textbf{k=10} & \textbf{k=30} & \textbf{k=50} & \textbf{k=100} \\ \hline
    	\multirow{4}{*}{\textbf{\kone}} & \multirow{2}{*}{\textbf{Full}} & \textbf{Avg} & 79.79 & 87.48 & 90.04 & 92.97 & 79.99 & 88.23 & 91.16 & {\textbf{94.07}} \\
    	&  & \textbf{Pft} & 19.60 & 28.55 & 33.24 & 39.20 & 19.89 & 30.40 & 34.52 & {\textbf{40.91}} \\
    	& \multirow{2}{*}{\textbf{Half}} & \textbf{Avg} & 94.58 & 97.58 & 98.48 & 99.02 & 95.27 & 98.30 & 99.20 & {\textbf{99.62}} \\
    	&  & \textbf{Pft} & 67.88 & 76.36 & 80.30 & 82.73 & 68.48 & 78.79 & 81.52 & {\textbf{87.27}} \\ \hline
    	\multirow{4}{*}{\textbf{\kocl}} & \multirow{2}{*}{\textbf{Full}} & \textbf{Avg} & 96.90 & 98.73 & 98.93 & 99.33 & 97.20 & 98.82 & 99.42 & {\textbf{99.77}} \\
    	&  & \textbf{Pft} & 69.85 & 87.17 & 91.81 & 94.94 & 72.50 & 90.49 & 93.94 & {\textbf{95.79}} \\
    	& \multirow{2}{*}{\textbf{Half}} & \textbf{Avg} & 97.65 & 99.20 & 99.44 & 99.73 & 97.68 & 99.27 & 99.50 & {\textbf{99.80}} \\
    	&  & \textbf{Pft} & 79.10 & 85.97 & 88.36 & 91.04 & 79.70 & 86.57 & 89.25 & {\textbf{91.64}} \\ \hline
    	\multirow{4}{*}{\textbf{\kbone}} & \multirow{2}{*}{\textbf{Full}} & \textbf{Avg} & 90.04 & 95.39 & 96.65 & 97.92 & 90.12 & 96.22 & 97.84 & {\textbf{98.71}} \\
    	&  & \textbf{Pft} & 34.90 & 47.45 & 54.31 & 65.10 & 36.47 & 52.94 & 60.00 & {\textbf{71.37}} \\
    	& \multirow{2}{*}{\textbf{Half}} & \textbf{Avg} & 92.05 & 96.83 & 97.33 & 98.63 & 92.17 & 97.39 & 99.19 & {\textbf{99.50}} \\
    	&  & \textbf{Pft} & 40.37 & 57.76 & 62.73 & 67.08 & 44.10 & 62.11 & 65.84 & {\textbf{73.29}} \\ \hline
    	\multirow{4}{*}{\textbf{\kbthree}} & \multirow{2}{*}{\textbf{Full}} & \textbf{Avg} & 94.29 & 98.10 & 99.29 & {\textbf{99.52}} & - & - & - & - \\
    	&  & \textbf{Pft} & 58.57 & 81.43 & 87.14 & {\textbf{91.43}} & - & - & - & - \\
    	& \multirow{2}{*}{\textbf{Half}} & \textbf{Avg} & 96..67 & 99.63 & 100.00 & {\textbf{100.00}} & - & - & - & - \\
    	&  & \textbf{Pft} & 77.78 & 88.89 & 91.11 & {\textbf{95.56}} & - & - & - & - \\ \hline
    	\multicolumn{11}{c}{\footnotesize Hybrid-2: Hybrid Layered Decoder Model, OPCS-BS: Output Parameter Constraint Satisfaction fused Beam Search} \\
    	\multicolumn{11}{c}{\footnotesize Avg: Average Accuracy, Pft: Perfect Prediction}
    \end{tabular}
\end{table*}

The RNN based model used here has various hyperparameters, which have a considerable impact on prediction accuracy. Key findings from trying various combinations for different models presented in Section \ref{sec:dev} are as follows: 

\begin{itemize}
    \item The size of the internal state vector $|e|$ for the encoder-decoder architecture strongly influences the quality of results, $|e|=256$ gives the best results. Moreover, setting decoder LSTM \emph{Dropout} and \emph{Recurrent Dropout} to 0.2 (i.e., 20\%) improved test results by preventing over-fitting.
    \item The tunable hyperparameters of both attention and modified attention models include the number of input/output layers, cell sizes of the pre-attention bi-LSTM $n_a$, post-attention LSTM $n_s$, and the number of nodes in the Dense layer $n_d$. $n_a=256$, $n_s=512$, and $n_d=2$ give the best results.
    \item In addition to the number of Conv and Sequential layers, tunable hyperparameters for the Hybrid Model include those specifically related to Conv1D as well as bi-LSTM. Hyperparameters of Conv1D include the number of filters $f$, the size of each filter, also known as kernel size $k$, and the convolutional stride $s$. For the bi-LSTM, hyperparameters include cell size of each unit, dropout, and recurrent dropout. A cell size of \emph{256} for most of the kernels gives the best results.
\end{itemize}

\subsection{Training convergence analysis}
\label{results:learning}
\begin{figure}[ht]
	\centering
	\includegraphics[clip=true, width=0.47\textwidth]{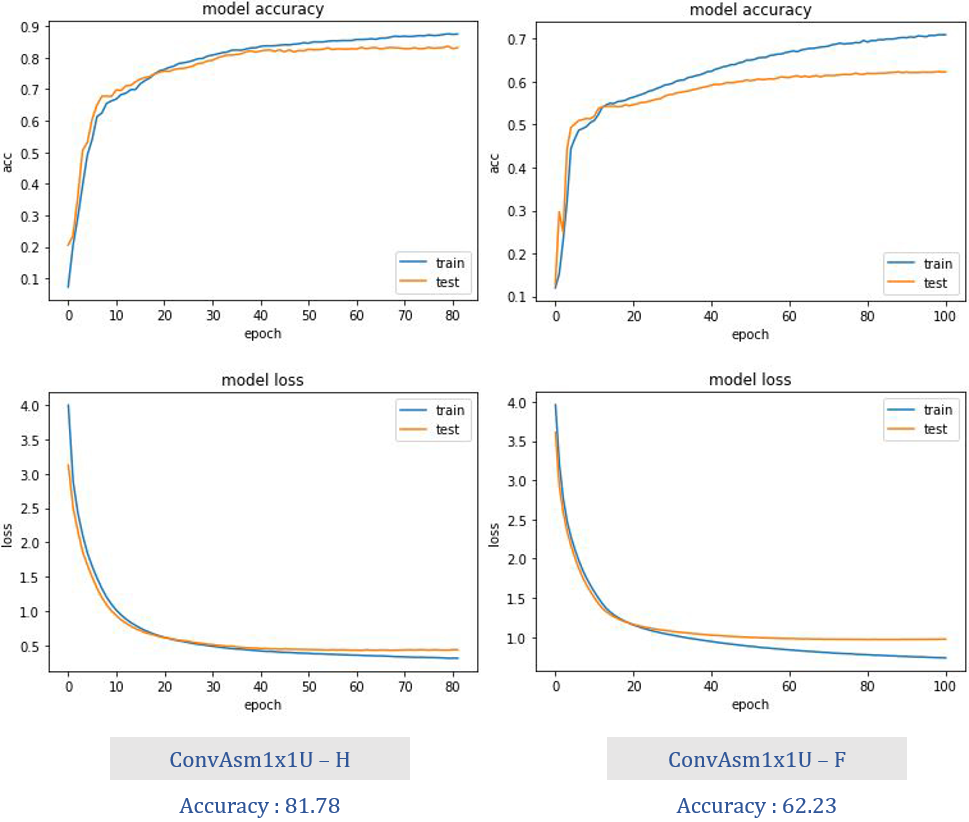}
	\caption{Change in Accuracy and Loss w.r.t Number of Epochs for {\kone} Half and Full Precision}
	\label{fig:5-4_k-1}
\end{figure}

\begin{figure}[ht]
	\centering
	\includegraphics[clip=true, width=0.47\textwidth]{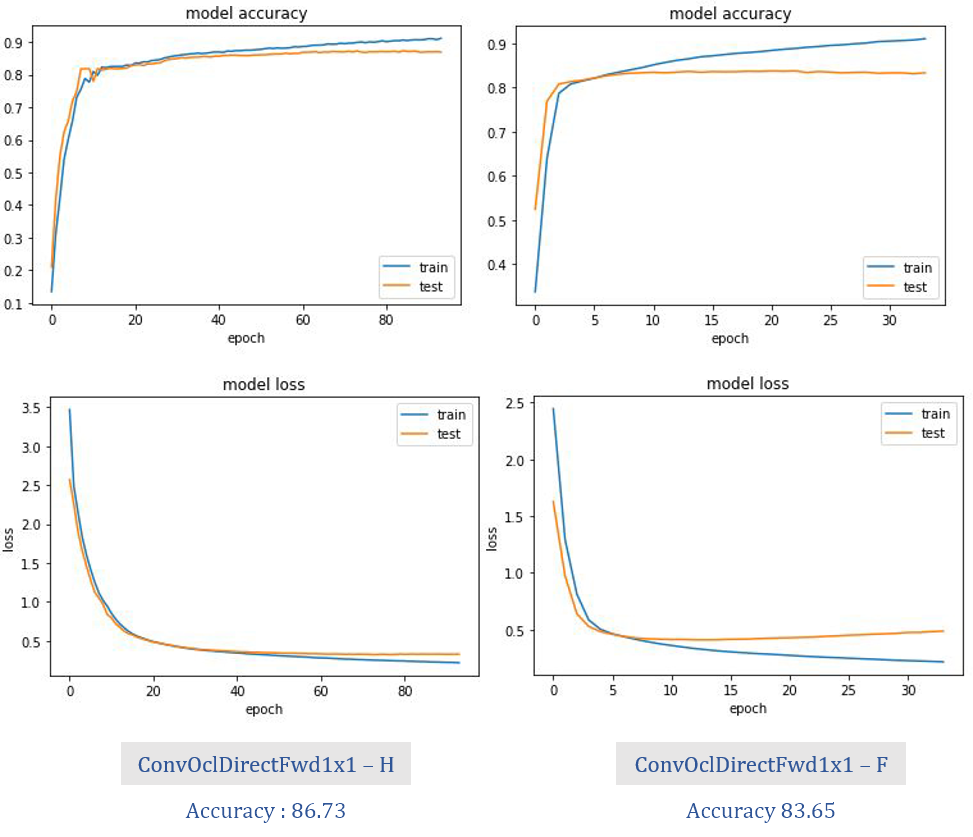}
	\caption{Change in Accuracy and Loss w.r.t Number of Epochs for {\kocl} Half and Full Precision}
	\label{fig:5-4_k-4}
\end{figure}

\begin{figure}[ht]
	\centering
	\includegraphics[clip=true, width=0.47\textwidth]{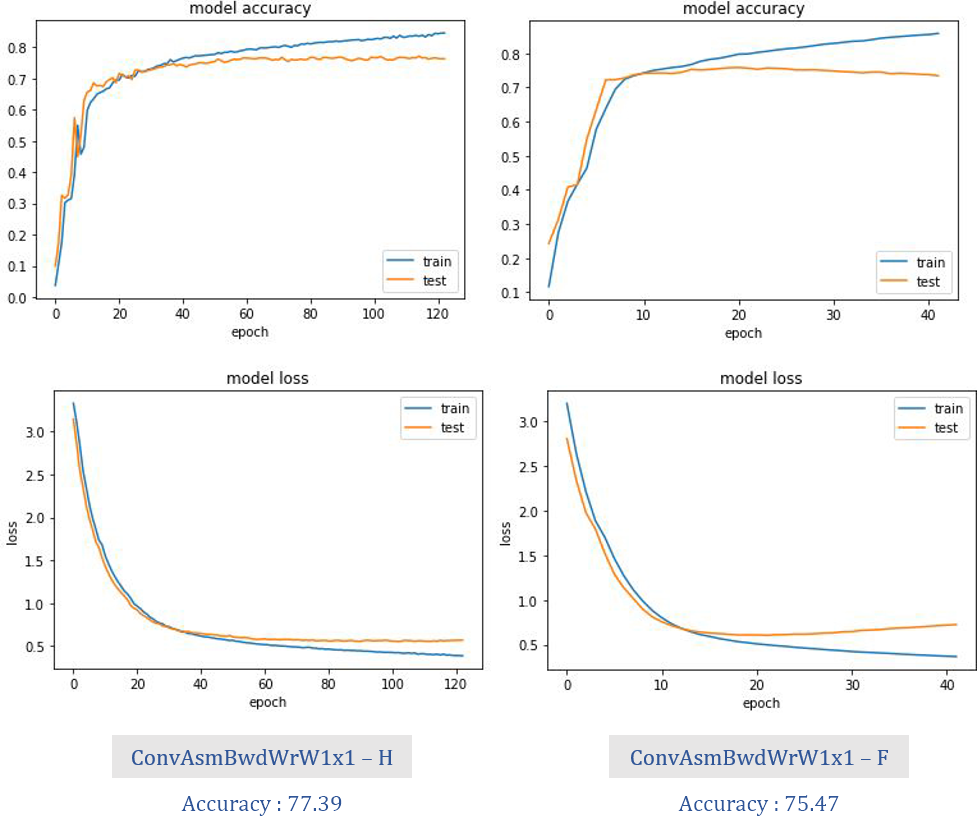}
	\caption{Change in Accuracy and Loss w.r.t Number of Epochs for {\kbone} Half and Full Precision}
	\label{fig:5-4_k-2}
\end{figure}

\begin{figure}[ht]
	\centering
	\includegraphics[clip=true, width=0.47\textwidth]{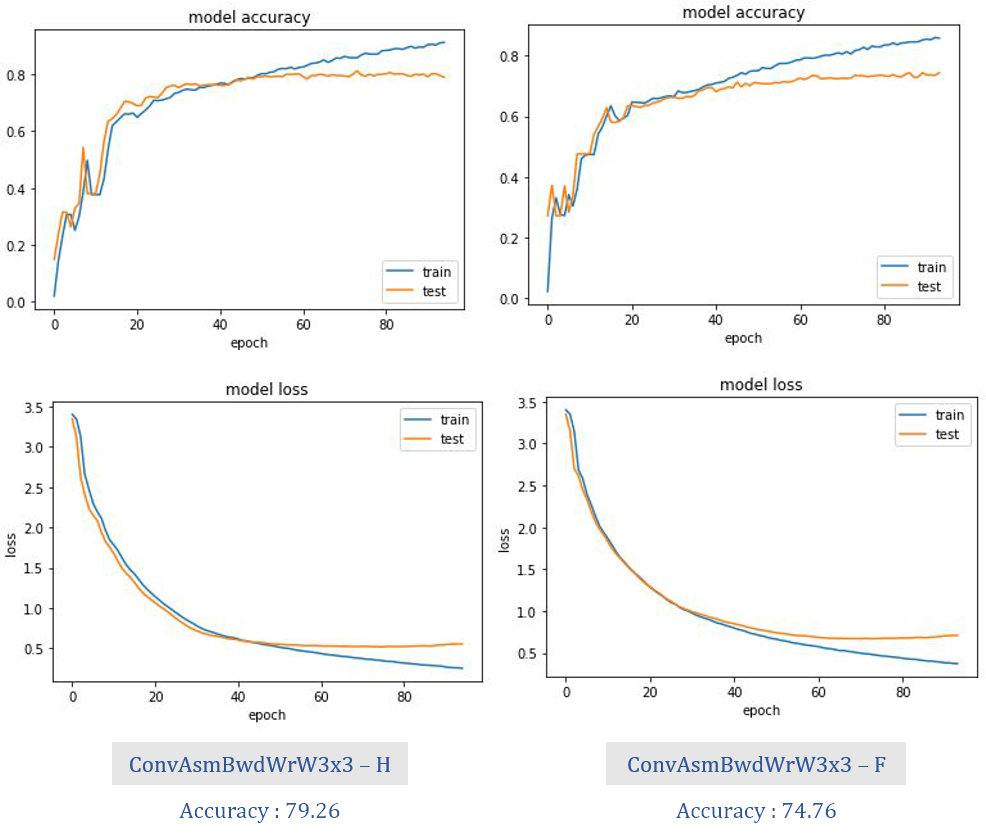}
	\caption{Change in Accuracy and Loss w.r.t Number of Epochs for {\kbthree} Half and Full Precision}
	\label{fig:5-4_k-3}
\end{figure}

Analysis of train-test accuracy improvement (and correspondingly loss reduction) with the increasing number of epochs gives insight into the learning of the Hybrid {\sts} model on various kernels. The change in accuracy and loss as a function of the number of epochs is shown for all four Kernels in Figures~\ref{fig:5-4_k-1},\ref{fig:5-4_k-4},\ref{fig:5-4_k-2},\ref{fig:5-4_k-3}. {\kone} and {\kocl} kernels in half-precision mode display closest training and test accuracy improvement (and corresponding loss reduction) with successive epochs. This learning performance corresponds to the comparatively better results seen for these kernels. Overall, additional data would aid learning for certain full-precision kernels.

\section{Conclusion}
\label{sec:concl}
\emph{\sts} models are powerful ML constructs that perform well in many domains. In this work, {\sts} models in the presence of limited data samples have shown the capacity to learn a discrete and complex search space with more than 95\% accuracy for many kernels. Convolutional (Conv) Layers are most suitable for forming data abstractions and representations, especially for problems such as {\kpp} where the sequences are not related in the temporal domain but have dependencies that originate from complex functions with multiple dimensions. This work indicates that it is possible to learn the computing behavior of a GPU using an RNN and use this model to predict the optimal kernel tuning parameters for compute kernels. 

\ifCLASSOPTIONcaptionsoff
  \newpage
\fi

\bibliographystyle{IEEEtran}
\bibliography{seqkernel}


%

\end{document}